\documentclass[10pt,twocolumn,letterpaper]{article}

\usepackage{iccv}
\usepackage{times}
\usepackage{epsfig}
\usepackage{graphicx}
\usepackage{amsmath}
\usepackage{amssymb}

\makeatletter
\renewcommand{\paragraph}{%
  \@startsection{paragraph}{4}%
  {\z@}{1.625ex \@plus 1ex \@minus .2ex}{-1em}%
  {\normalfont\normalsize\bfseries}%
}
\makeatother

\usepackage{booktabs}
\usepackage{tabularx}
\newcolumntype{Y}{>{\centering\arraybackslash}X}
\newcolumntype{P}[1]{>{\centering\arraybackslash}p{#1}}
\newcolumntype{M}[1]{>{\centering\arraybackslash}m{#1}}

\usepackage{subcaption}
\DeclareMathOperator{\msa}{MSA}
\DeclareMathOperator{\lnorm}{LN}
\DeclareMathOperator{\mlp}{MLP}
\DeclareMathOperator{\softmax}{Softmax}
\DeclareMathOperator{\attention}{Attention}
\DeclareMathOperator{\concat}{Concat}

\usepackage{mathtools}

\DeclarePairedDelimiter\floor{\lfloor}{\rfloor}

\usepackage[pagebackref=true,breaklinks=true,letterpaper=true,colorlinks,bookmarks=false]{hyperref}

\iccvfinalcopy %

\def\iccvPaperID{2188} %

\ificcvfinal\fi

\begin{document}

\title{\vspace{-\baselineskip} ViViT: A Video Vision Transformer}

\author{Anurag Arnab\thanks{Equal contribution} \quad Mostafa Dehghani\footnotemark[1] \quad Georg Heigold \quad Chen Sun \quad Mario Lučić\thanks{Equal advising}  \quad  Cordelia Schmid\footnotemark[2] \\
Google Research \\
{\tt\small \{aarnab, dehghani, heigold, chensun, lucic, cordelias\}@google.com}
}

\maketitle
\ificcvfinal\thispagestyle{empty}\fi

\begin{abstract}
	
We present pure-transformer based models for video classification, drawing upon the recent success of such models in image classification.
Our model extracts spatio-temporal tokens from the input video, which are then encoded by a series of transformer layers.
In order to handle the long sequences of tokens encountered in video, we propose several, efficient variants of our model which factorise the spatial- and temporal-dimensions of the input. %
Although transformer-based models are known to only be effective when large training datasets are available, we show how we can effectively regularise the model during training and leverage pretrained image models to be able to train on comparatively small datasets.
We conduct thorough ablation studies, and achieve state-of-the-art results on multiple video classification benchmarks including Kinetics 400 and 600, Epic Kitchens, Something-Something v2 and Moments in Time, outperforming prior methods based on deep 3D convolutional networks.
To facilitate further research, we release code at \href{https://github.com/google-research/scenic/tree/main/scenic/projects/vivit}{https://github.com/google-research/scenic}.

\end{abstract}

\section{Introduction}

Approaches based on deep convolutional neural networks have advanced the state-of-the-art across many standard datasets for vision problems since AlexNet~\cite{krizhevsky_neurips_2012}.
At the same time, the most prominent architecture of choice in sequence-to-sequence modelling (e.g. in natural language processing) is the transformer~\cite{vaswani_neurips_2017}, which does not use convolutions, but is based on multi-headed self-attention.
This operation is particularly effective at modelling long-range dependencies and allows the model to attend over all elements in the input sequence.
This is in stark contrast to convolutions where the corresponding ``receptive field'' is limited, and grows linearly with the depth of the network.%

The success of attention-based models in NLP has recently inspired approaches in computer vision to integrate transformers into CNNs~\cite{wang_cvpr_2018, carion_eccv_2020}, as well as some attempts to replace convolutions completely~\cite{parmar_icml_2018, bello_iccv_2019, ramachandran_neurips_2019}. 
However, it is only very recently with the Vision Transformer (ViT)~\cite{dosovitskiy_iclr_2021}, that a pure-transformer based architecture has outperformed its convolutional counterparts in image classification.
Dosovitskiy~\etal~\cite{dosovitskiy_iclr_2021} closely followed the original transformer architecture of~\cite{vaswani_neurips_2017}, and noticed that its main benefits were observed at large scale -- as transformers lack some of the inductive biases of convolutions (such as translational equivariance), they seem to require more data~\cite{dosovitskiy_iclr_2021} or stronger regularisation~\cite{touvron_arxiv_2020}.

Inspired by ViT, and the fact that attention-based architectures are an intuitive choice for modelling long-range contextual relationships in video, we develop several transformer-based models for video classification.
Currently, the most performant models are based on deep 3D convolutional architectures~\cite{carreira_cvpr_2017, feichtenhofer_cvpr_2020, feichtenhofer_iccv_2019} which were a natural extension of image classification CNNs~\cite{he_cvpr_2016, szegedy_cvpr_2015}.
Recently, these models were augmented by incorporating self-attention into their later layers to better capture long-range dependencies~\cite{wang_cvpr_2018, girdhar_cvpr_2019, wu_cvpr_2019, arnab_graph_structured_iccv_2021}.

\begin{figure*}
    \centering
    \includegraphics[width=1.0\linewidth]{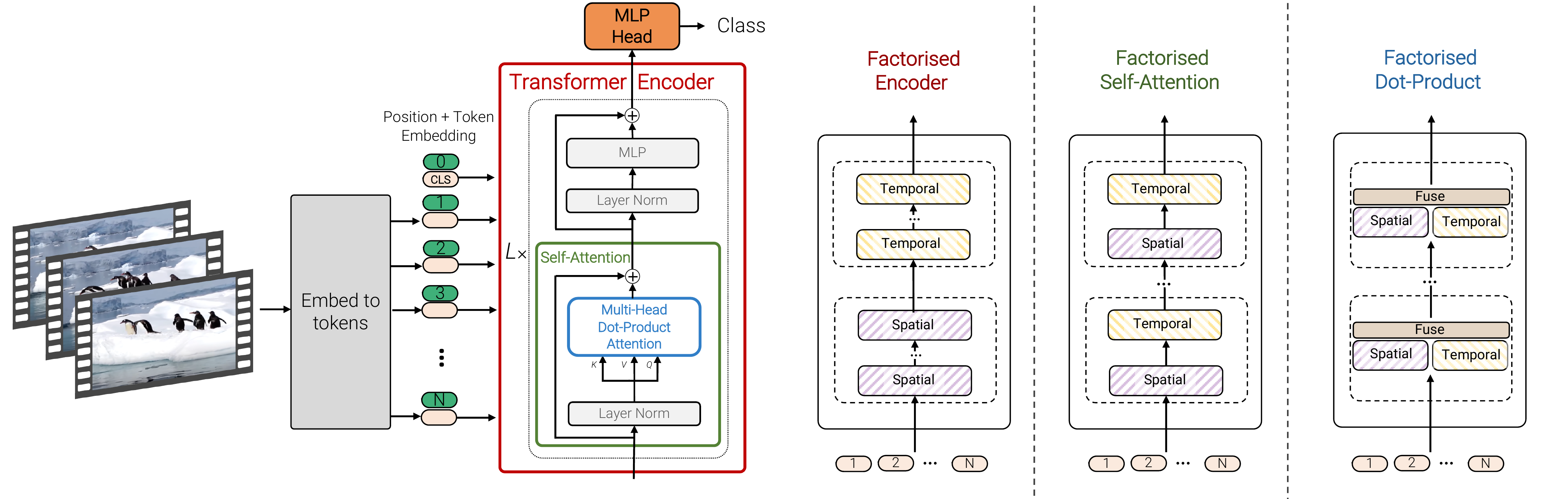}
	\caption{
		We propose a pure-transformer architecture for video classification, inspired by the recent success of such models for images~\cite{dosovitskiy_iclr_2021}.
		To effectively process a large number of spatio-temporal tokens, we develop several model variants which factorise different components of the transformer encoder over the spatial- and temporal-dimensions.
		As shown on the right, these factorisations correspond to different attention patterns over space and time.
	}
	\vspace{-3mm}
    \label{fig:teaser}
\end{figure*}

As shown in Fig.~\ref{fig:teaser}, we propose pure-transformer models for video classification.
The main operation performed in this architecture is self-attention, and it is computed on a sequence of spatio-temporal tokens that we extract from the input video.
To effectively process the large number of spatio-temporal tokens that may be encountered in video, we present several methods of factorising our model along spatial and temporal dimensions to increase efficiency and scalability.
Furthermore, to train our model effectively on smaller datasets, we show how to reguliarise our model during training and leverage pretrained image models.

We also note that convolutional models have been developed by the community for several years, and there are thus many ``best practices'' associated with such models.
As pure-transformer models present different characteristics, we need to determine the best design choices for such architectures. 
We conduct a thorough ablation analysis of tokenisation strategies, model architecture and regularisation methods.
Informed by this analysis, we achieve state-of-the-art results on multiple standard video classification benchmarks, including Kinetics 400 and 600~\cite{kay_arxiv_2017}, Epic Kitchens 100~\cite{damen_arxiv_2020}, Something-Something v2~\cite{goyal_iccv_2017} and Moments in Time~\cite{monfort_pami_2019}.

\section{Related Work}

Architectures for video understanding have mirrored advances in image recognition.
Early video research used hand-crafted features to encode appearance and motion information~\cite{laptev_ijcv_2005, wang_ijcv_2013}.
The success of AlexNet on ImageNet~\cite{krizhevsky_neurips_2012, deng_cvpr_2009} initially led to the repurposing of 2D image convolutional networks (CNNs) for video as ``two-stream'' networks~\cite{karpathy_cvpr_2014, simonyan_neurips_2014, ng_cvpr_2015}.
These models processed RGB frames and optical flow images independently before fusing them at the end.
Availability of larger video classification datasets such as Kinetics~\cite{kay_arxiv_2017} subsequently facilitated the training of spatio-temporal 3D CNNs~\cite{carreira_cvpr_2017, feichtenhofer_neurips_2016,tran_iccv_2015} which have significantly more parameters and thus require larger training datasets.
As 3D convolutional networks require significantly more computation than their image counterparts, many architectures factorise convolutions across spatial and temporal dimensions and/or use grouped convolutions~\cite{sun_iccv_2015,tran_iccv_2019,tran_cvpr_2018,xie_s3d_eccv_2018, feichtenhofer_cvpr_2020}.
We also leverage factorisation of the spatial and temporal dimensions of videos to increase efficiency, but in the context of transformer-based models.

Concurrently, in natural language processing (NLP), Vaswani~\etal~\cite{vaswani_neurips_2017} achieved state-of-the-art results by replacing convolutions and recurrent networks with the transformer network that consisted only of self-attention, layer normalisation and multilayer perceptron (MLP) operations.
Current state-of-the-art architectures in NLP~\cite{devlin_naacl_2019, raffel_jmlr_2020} remain transformer-based, and have been scaled to web-scale datasets~\cite{brown_gpt3_neurips_2020}.
Many variants of the transformer have also been proposed to reduce the computational cost of self-attention when processing longer sequences~\cite{child_arxiv_2019, choromanski_arxiv_2020,kitaev_iclr_2020, tay2020long, tay2020efficient, wang_arxiv_2020} and to improve parameter efficiency~\cite{lan2019albert, dehghani2018universal}.
Although self-attention has been employed extensively in computer vision, it has, in contrast, been typically incorporated as a layer at the end or in the later stages of the network~\cite{wang_cvpr_2018, carion_eccv_2020,  huang_ccnet_iccv_2019,wang_vistr_arxiv_2020, zhang_dgmn_cvpr_2020} or to augment residual blocks~\cite{hu_squeeze_excite_cvpr_2018, cao_gcnet_cvprw_2019, chen_danet_neurips_2018, srinivas_cvpr_2021} within a ResNet architecture~\cite{he_cvpr_2016}.

Although previous works attempted to replace convolutions in vision architectures~\cite{parmar_icml_2018, ramachandran_neurips_2019, shen_gsa_iclr_2021}, it is only very recently that Dosovitisky~\etal~\cite{dosovitskiy_iclr_2021} showed with their ViT architecture that pure-transformer networks, similar to those employed in NLP, can achieve state-of-the-art results for image classification too.
The authors showed that such models are only effective at large scale, as transformers lack some of inductive biases of convolutional networks (such as translational equivariance), and thus require datasets larger than the common ImageNet ILSRVC dataset~\cite{deng_cvpr_2009} to train.
ViT has inspired a large amount of follow-up work in the community, and we note that there are a number of concurrent approaches on extending it to other tasks in computer vision~\cite{wang_maxdeeplab_arxiv_2020,wang_pvt_arxiv_2021, zhao_point_transformer_arxiv_2020,zheng_setr_arxiv_2020} and improving its data-efficiency~\cite{touvron_arxiv_2020,pan_hvt_arxiv_2021}.
In particular, \cite{bertasius_arxiv_2021,neimark_arxiv_2021} have also proposed transformer-based models for video.

In this paper, we develop pure-transformer architectures for video classification.
We propose several variants of our model, including those that are more efficient by factorising the spatial and temporal dimensions of the input video.
We also show how additional regularisation and pretrained models can be used to combat the fact that video datasets are not as large as their image counterparts that ViT was originally trained on.
Furthermore, we outperform the state-of-the-art across five popular datasets.

\section{Video Vision Transformers}
\label{sec:method}

We start by summarising the recently proposed Vision Transformer~\cite{dosovitskiy_iclr_2021} in~Sec.~\ref{sec:method_background}, and then discuss two approaches for extracting tokens from video in~Sec.~\ref{sec:method_input_encoding}. Finally, we develop several transformer-based architectures for video classification in Sec.~\ref{sec:method_video_models} and~\ref{sec:method_initialisation}.

\subsection{Overview of Vision Transformers (ViT)}
\label{sec:method_background}

Vision Transformer (ViT)~\cite{dosovitskiy_iclr_2021} adapts the transformer architecture of~\cite{vaswani_neurips_2017} to process 2D images with minimal changes.
In particular, ViT extracts $N$ non-overlapping image patches, $x_i \in \mathbb{R}^{h \times w}$, %
performs a linear projection and then rasterises them into 1D tokens $z_i \in \mathbb{R}^d$. The sequence of tokens input to the following transformer encoder is
\begin{equation}
\mathbf{z} = [z_{cls}, \mathbf{E}x_1, \mathbf{E}x_2, \ldots , \mathbf{E}x_N] + \mathbf{p},
\label{eq:tokens}
\end{equation}
where the projection by $\mathbf{E}$ is equivalent to a 2D convolution.
As shown in Fig.~\ref{fig:teaser}, an optional learned classification token $z_{cls}$ is prepended to this sequence, and its representation at the final layer of the encoder serves as the final representation used by the classification layer~\cite{devlin_naacl_2019}.
In addition, a learned positional embedding, $\mathbf{p} \in \mathbb{R}^{N \times d}$, is added to the tokens to retain positional information, as the subsequent self-attention operations in the transformer are permutation invariant.
The tokens are then passed through an encoder consisting of a sequence of $L$ transformer layers.
Each layer $\ell$ comprises of Multi-Headed Self-Attention~\cite{vaswani_neurips_2017}, layer normalisation (LN)~\cite{ba_arxiv_2016}, and MLP blocks as follows:
\begin{align}
\mathbf{y}^{\ell} &= \text{MSA}(\text{LN}(\mathbf{z}^\ell)) +\mathbf{z}^\ell  \\
\mathbf{z}^{\ell + 1} &= \text{MLP}(\text{LN}(\mathbf{y}^\ell)) + \mathbf{y}^\ell.
\end{align}
The MLP consists of two linear projections separated by a GELU non-linearity~\cite{hendrycks_arxiv_2016} and the token-dimensionality, $d$, remains fixed throughout all layers.
Finally, a linear classifier is used to classify the encoded input based on $z_{cls}^L \in \mathbb{R}^d$, if it was prepended to the input, or a global average pooling of all the tokens, $\mathbf{z}^{L}$, otherwise.

As the transformer~\cite{vaswani_neurips_2017}, which forms the basis of ViT~\cite{dosovitskiy_iclr_2021}, is a flexible architecture that can operate on any sequence of input tokens $\mathbf{z} \in \mathbb{R}^{N \times d}$, we describe strategies for tokenising videos next.

\subsection{Embedding video clips}
\label{sec:method_input_encoding}
We consider two simple %
methods for mapping a video $\mathbf{V} \in \mathbb{R}^{T \times H \times W \times C}$ to a sequence of tokens $\mathbf{\tilde{z}} \in \mathbb{R}^{n_t \times n_h \times n_w \times d}$.
We then add the positional embedding and reshape into $\mathbb{R}^{N \times d}$ to obtain $\mathbf{z}$, the input to the transformer.%

\paragraph{Uniform frame sampling} 

\begin{figure}
    \centering
     \includegraphics[width=1.0\linewidth]{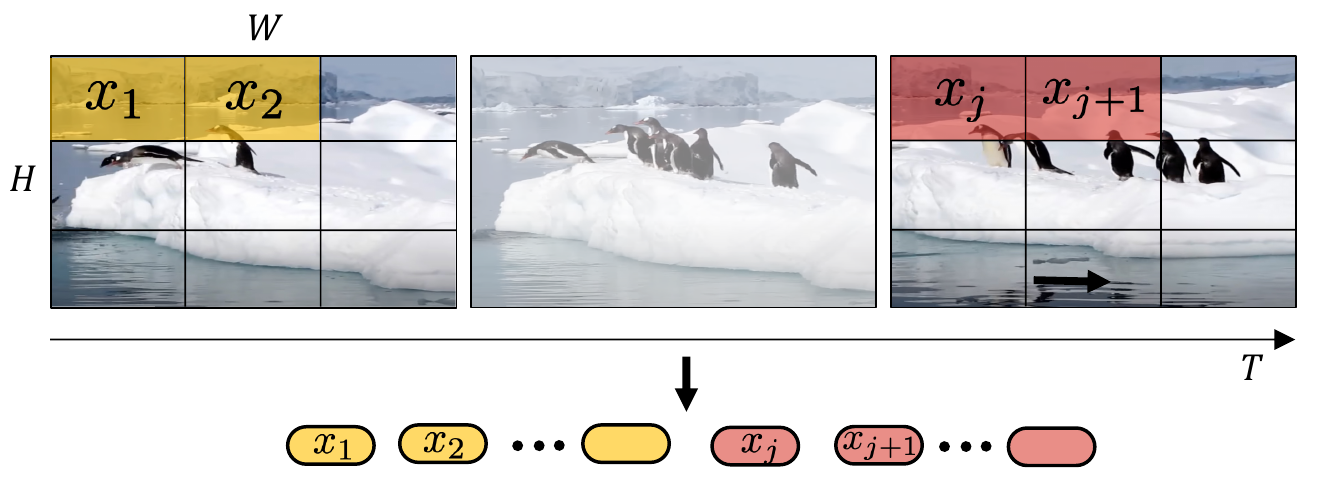}
    \caption{Uniform frame sampling: We simply sample $n_t$ frames, and embed each 2D frame independently following ViT~\cite{dosovitskiy_iclr_2021}.}
    \label{fig:tokenizer_2d}
\end{figure}

As illustrated in Fig.~\ref{fig:tokenizer_2d}, a straightforward method of tokenising the input video is to uniformly sample $n_t$ frames from the input video clip, embed each 2D frame independently using the same method as ViT~\cite{dosovitskiy_iclr_2021}, and concatenate all these tokens together.
Concretely, if $n_h \cdot n_w$ non-overlapping image patches are extracted from each frame, as in~\cite{dosovitskiy_iclr_2021}, then a total of $n_t \cdot n_h \cdot n_w$ tokens will be forwarded through the transformer encoder. 
Intuitively, this process may be seen as simply constructing a large 2D image to be tokenised following ViT.
We note that this is the input embedding method employed by the concurrent work of~\cite{bertasius_arxiv_2021}.

\paragraph{Tubelet embedding}
\begin{figure}
    \centering
    \includegraphics[width=0.9\linewidth]{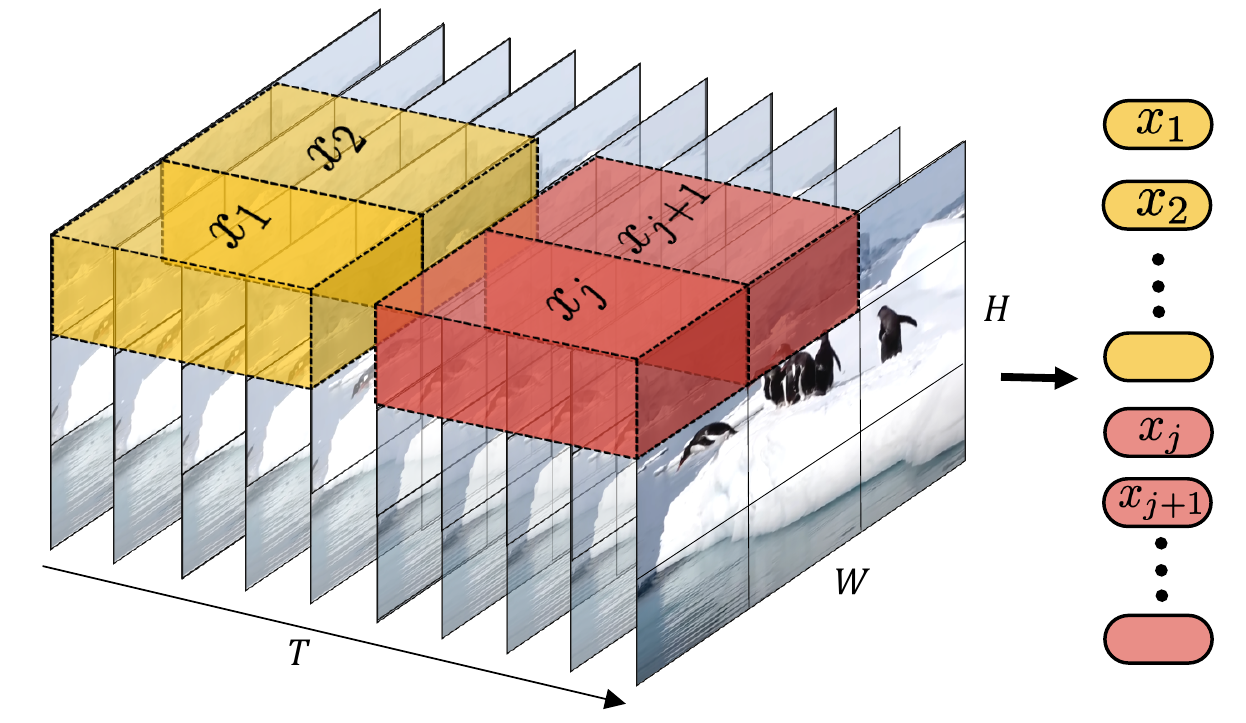}
    \caption{Tubelet embedding. We extract and linearly embed non-overlapping tubelets that span the spatio-temporal input volume. %
    }
    \vspace{-3mm}
    \label{fig:tokenizer_3d}
\end{figure}
An alternate method, as shown in Fig.~\ref{fig:tokenizer_3d}, is to extract non-overlapping, spatio-temporal ``tubes'' from the input volume, and to linearly project this to $\mathbb{R}^d$.
This method is an extension of ViT's embedding to 3D, and corresponds to a 3D convolution.
For a tubelet of dimension $t \times h \times w$, $n_t = \floor{\frac{T}{t}}$, $n_h = \floor{\frac{H}{h}}$ and $n_w = \floor{\frac{W}{w}}$, tokens are extracted from the temporal, height, and width dimensions respectively.
Smaller tubelet dimensions thus result in more tokens which increases the computation. Intuitively, this method fuses spatio-temporal information during tokenisation, in contrast to ``Uniform frame sampling'' where temporal information from different frames is fused by the transformer.

\subsection{Transformer Models for Video}
\label{sec:method_video_models}

As illustrated in Fig.~\ref{fig:teaser}, we propose multiple transformer-based architectures.
We begin with a straightforward extension of ViT~\cite{dosovitskiy_iclr_2021} that models pairwise interactions between all spatio-temporal tokens, and then develop more efficient variants which factorise the spatial and temporal dimensions of the input video at various levels of the transformer architecture.

\paragraph{Model 1: Spatio-temporal attention}
This model simply forwards all spatio-temporal tokens extracted from the video, $\mathbf{z}^{0}$, through the transformer encoder.
We note that this has also been explored concurrently by~\cite{bertasius_arxiv_2021} in their ``Joint Space-Time'' model.
In contrast to CNN architectures, where the receptive field grows linearly with the number of layers, each transformer layer models all pairwise interactions between all spatio-temporal tokens, and it thus models long-range interactions across the video from the first layer.
However, as it models all pairwise interactions, Multi-Headed Self Attention (MSA)~\cite{vaswani_neurips_2017} has quadratic complexity with respect to the number of tokens.
This complexity is pertinent for video, as the number of tokens increases linearly with the number of input frames, and motivates the development of more efficient architectures next.%

\paragraph{Model 2: Factorised encoder}
\begin{figure}
    \vspace{-0.5\baselineskip}
    \centering
    \includegraphics[width=1.0\linewidth]{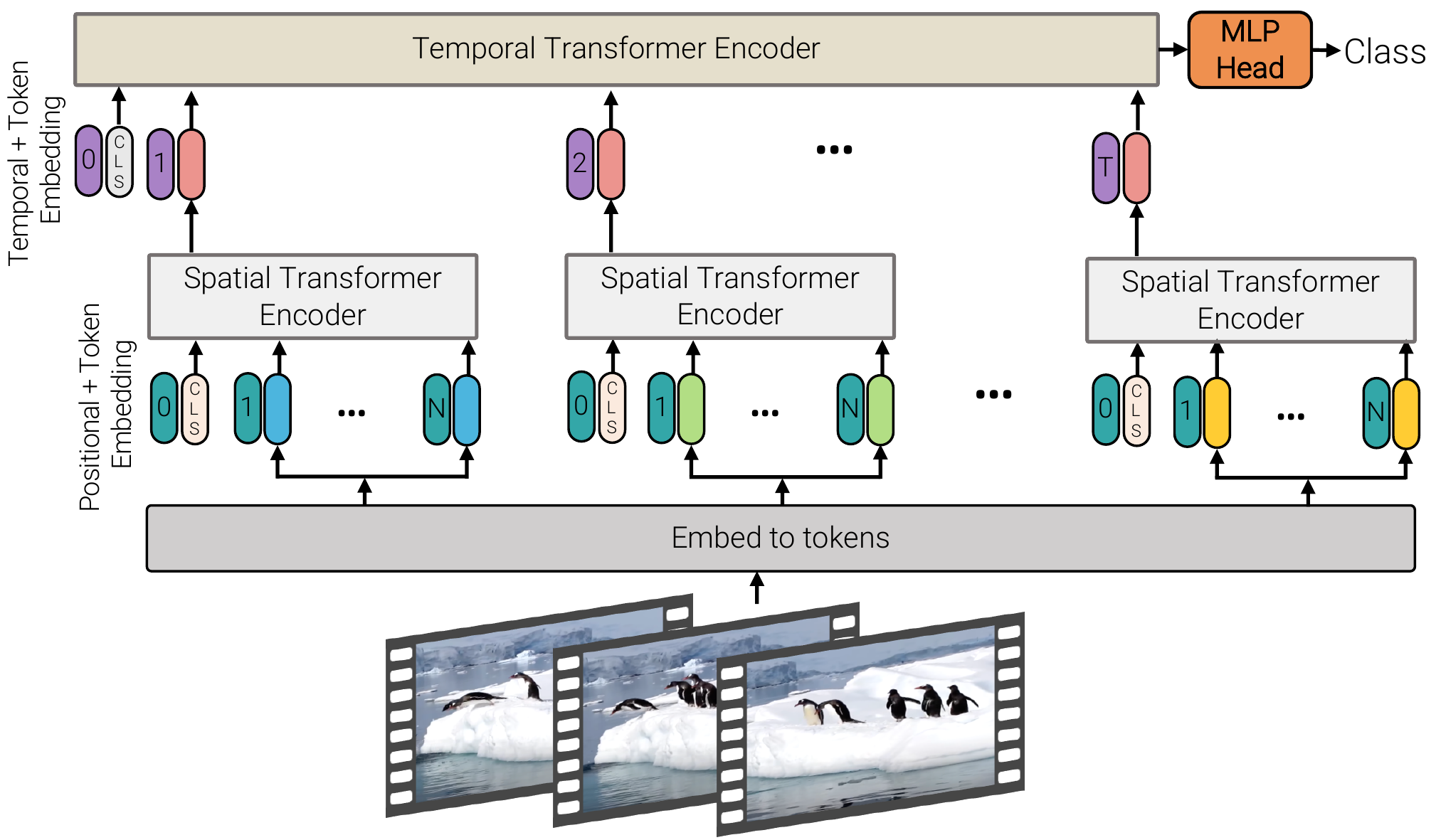}
    \vspace{0.1em}
    \caption{
    	Factorised encoder (Model 2). 
    	This model consists of two transformer encoders in series: the first models interactions between tokens extracted from the same temporal index to produce a latent representation per time-index.
    	The second transformer models interactions between time steps.
    	It thus corresponds to a ``late fusion'' of spatial- and temporal information.
    }
    \vspace{-\baselineskip}
    \label{fig:factorised_encoder}
\end{figure}

As shown in Fig.~\ref{fig:factorised_encoder}, this model consists of two separate transformer encoders.
The first, spatial encoder, only models interactions between tokens extracted from the same temporal index. %
A representation for each temporal index, $h_i \in \mathbb{R}^d$, is obtained after $L_s$ layers: This is the encoded classification token, $z_{cls}^{L_s}$ if it was prepended to the input (Eq.~\ref{eq:tokens}), or a global average pooling from the tokens output by the spatial encoder, $\mathbf{z}^{L_s}$, otherwise.
The frame-level representations, $h_i$, are concatenated into $\mathbf{H} \in \mathbb{R}^{n_t \times d}$, and then forwarded through a temporal encoder consisting of $L_t$ transformer layers to model interactions between tokens from different temporal indices.
The output token of this encoder is then finally classified.

This architecture corresponds to a ``late fusion''~\cite{karpathy_cvpr_2014, simonyan_neurips_2014, wang_tsn_eccv_2016, neimark_arxiv_2021} of temporal information, and the initial spatial encoder is identical to the one used for image classification.
It is thus analogous to CNN architectures such as ~\cite{girdhar_neurips_2017, karpathy_cvpr_2014, wang_tsn_eccv_2016, zhou_trn_eccv_2018} which first extract per-frame features, and then aggregate them into a final representation before classifying them. %
Although this model has more transformer layers than Model 1 (and thus more parameters), it requires fewer floating point operations (FLOPs), as the two separate transformer blocks have a complexity of $\mathcal{O}({(n_h \cdot n_w)^2 + n_t^2)}$ compared to $\mathcal{O}((n_t \cdot n_h \cdot n_w)^2)$ of Model 1.

\paragraph{Model 3: Factorised self-attention}
\begin{figure}
    \vspace{-0.5\baselineskip}
    \centering
    \includegraphics[width=1.0\linewidth]{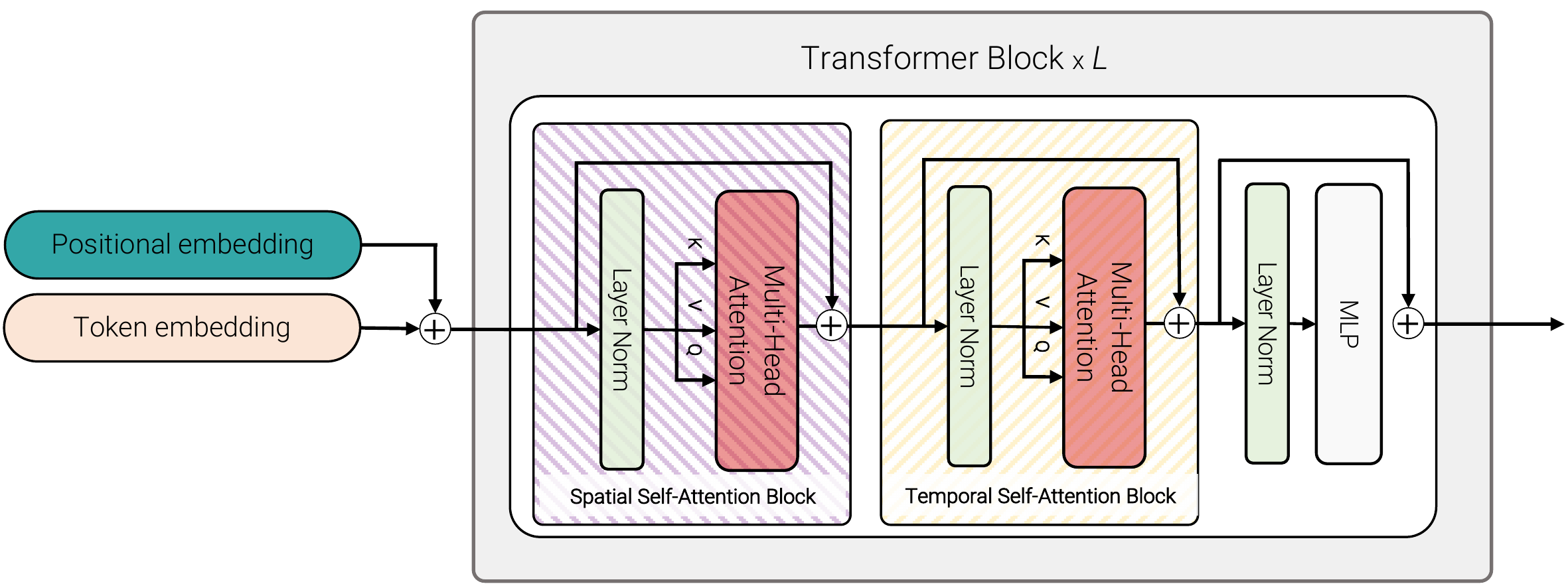}
        \vspace{0.1em}
        \caption{ 
    	Factorised self-attention (Model 3).
    	Within each transformer block, the multi-headed self-attention operation is factorised into two operations (indicated by striped boxes) that first only compute self-attention spatially, and then temporally.
    }
    \label{fig:factorised_self_attention}
    \vspace{-\baselineskip}
\end{figure}

This model, in contrast, contains the same number of transformer layers as Model 1.
However, instead of computing multi-headed self-attention across all pairs of tokens, $\mathbf{z}^{\ell}$, at layer $l$, we factorise the operation to first only compute self-attention spatially (among all tokens extracted from the same temporal index), and then temporally (among all tokens extracted from the same spatial index) as shown in Fig.~\ref{fig:factorised_self_attention}.
Each self-attention block in the transformer thus models spatio-temporal interactions, but does so more efficiently than Model 1 by factorising the operation over two smaller sets of elements, thus achieving the same computational complexity as Model 2. %
We note that factorising attention over input dimensions has also been explored in~\cite{ho_arxiv_2019, weissenborn_iclr_2020}, and concurrently in the context of video by~\cite{bertasius_arxiv_2021} in their ``Divided Space-Time'' model.

This operation can be performed efficiently by reshaping the tokens $\mathbf{z}$ from $\mathbb{R}^{1 \times n_t \cdot n_h \cdot n_w \cdot d}$ to $\mathbb{R}^{n_t \times n_h \cdot n_w \cdot d}$ (denoted by $\mathbf{z}_s$) to compute spatial self-attention.
Similarly, the input to temporal self-attention, $\mathbf{z}_t$ is reshaped to $\mathbb{R}^{n_h \cdot n_w \times  n_t \cdot d}$.
Here we assume the leading dimension is the ``batch dimension''.
Our factorised self-attention is defined as
\begin{align}
\mathbf{y}_{s}^{\ell}    &= \msa(\lnorm(\mathbf{z}^{\ell}_s)) + \mathbf{z}^{\ell}_s \\
\mathbf{y}_{t}^{\ell}    &= \msa(\lnorm(\mathbf{y}^{\ell}_s)) + \mathbf{y}^{\ell}_s  \label{eq:factorised_self_att_temporal} \\
\mathbf{z}^{\ell + 1}  &= \mlp(\lnorm(\mathbf{y}_t^{\ell})) + \mathbf{y}_t^\ell.
\end{align}
We observed that the order of spatial-then-temporal self-attention or temporal-then-spatial self-attention does not make a difference, provided that the model parameters are initialised as described in Sec.~\ref{sec:method_initialisation}.
Note that the number of parameters, however, increases compared to Model 1, as there is an additional self-attention layer (cf. Eq.~\ref{eq:selfattn}).
We do not use a classification token in this model, to avoid ambiguities when reshaping the input tokens between spatial and temporal dimensions.

\paragraph{Model 4: Factorised dot-product attention}
\begin{figure}
    \centering
    \includegraphics[width=1.0\linewidth]{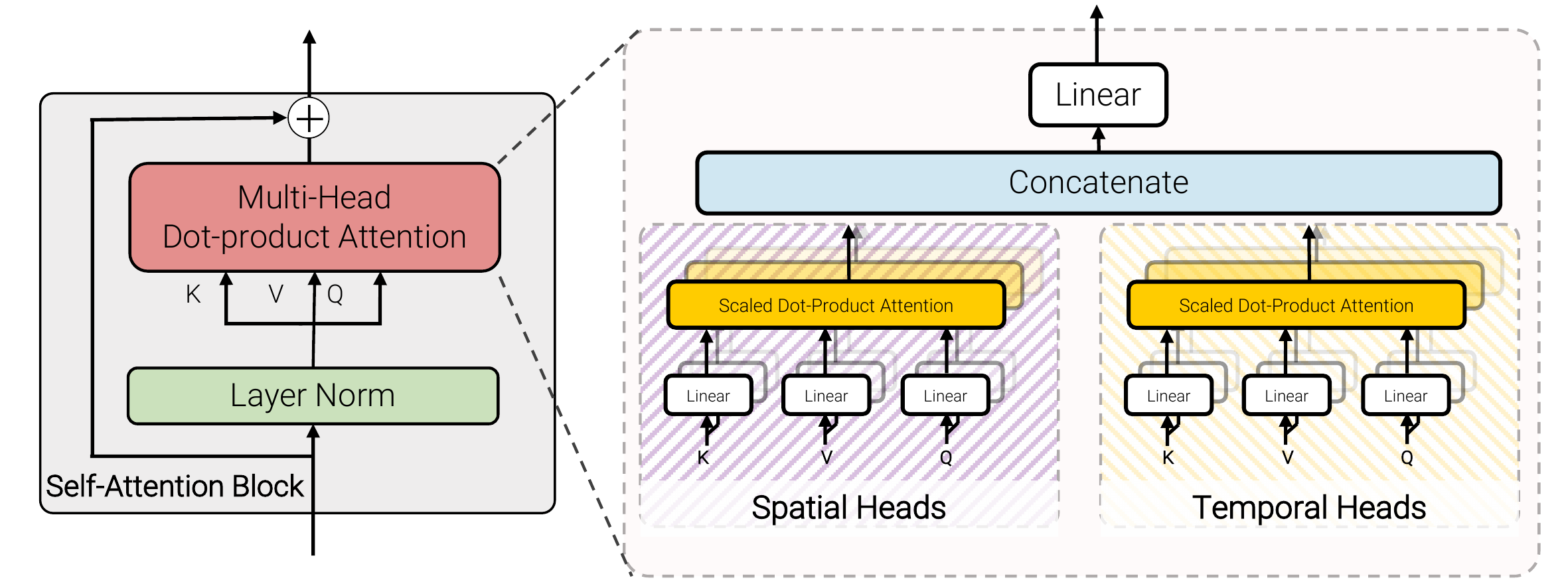}
    \caption{Factorised dot-product attention (Model 4).
    For half of the heads, we compute dot-product attention over only the spatial axes, and for the other half, over only the temporal axis.
    }
    \label{fig:factorised_dot_product_attention}
    \vspace{-\baselineskip}
\end{figure}
Finally, we develop a model which has the same computational complexity as Models 2 and 3, while retaining the same number of parameters as the unfactorised Model 1.
The factorisation of spatial- and temporal dimensions is similar in spirit to Model 3, but we factorise the multi-head dot-product attention operation instead (Fig.~\ref{fig:factorised_dot_product_attention}).
Concretely, we compute attention weights for each token separately over the spatial- and temporal-dimensions using different heads.
First, we note that the attention operation for each head is defined as
\begin{align}
\attention(\mathbf{Q}, \mathbf{K}, \mathbf{V}) = \softmax\left(\frac{\mathbf{Q} \mathbf{K}^\top}{\sqrt{d_k}} \right) \mathbf{V}. \label{eq:selfattn}
\end{align}
In self-attention, the queries $\mathbf{Q} = \mathbf{X} \mathbf{W}_q$, keys $\mathbf{K} = \mathbf{X} \mathbf{W}_k$, and values $\mathbf{V}= \mathbf{X} \mathbf{W}_v$ are linear projections of the input $\mathbf{X}$ with $\mathbf{X}, \mathbf{Q}, \mathbf{K}, \mathbf{V} \in \mathbb{R}^{N \times d}$. 
Note that in the unfactorised case (Model 1), the spatial and temporal dimensions are merged as $N = n_t \cdot n_h \cdot n_w$.

The main idea here is to modify the keys and values for each query to only attend over tokens from the same spatial- and temporal index by constructing $\mathbf{K}_s, \mathbf{V}_s \in \mathbb{R}^{n_h \cdot n_w \times d}$ and $\mathbf{K}_t, \mathbf{V}_t \in \mathbb{R}^{n_t \times d}$, namely the keys and values corresponding to these dimensions. Then, for half of the attention heads, we attend over tokens from the spatial dimension by computing $\mathbf{Y}_s = \attention(\mathbf{Q}, \mathbf{K}_s, \mathbf{V}_s)$, and for the rest we attend over the temporal dimension by computing $\mathbf{Y}_t = \attention(\mathbf{Q}, \mathbf{K}_t, \mathbf{V}_t)$.
Given that we are only changing the attention neighbourhood for each query, the attention operation has the same dimension as in the unfactorised case, namely $\mathbf{Y}_s, \mathbf{Y}_t \in \mathbb{R}^{N \times d}$. 
We then combine the outputs of multiple heads by concatenating them and using a linear projection~\cite{vaswani_neurips_2017}, 
$\mathbf{Y} = \concat(\mathbf{Y}_s, \mathbf{Y}_t) \mathbf{W}_O$.

\subsection{Initialisation by leveraging pretrained models}
\label{sec:method_initialisation}

ViT~\cite{dosovitskiy_iclr_2021} has been shown to only be effective when trained on large-scale datasets, as transformers lack some of the inductive biases of convolutional networks~\cite{dosovitskiy_iclr_2021}. %
However, even the largest video datasets such as Kinetics~\cite{kay_arxiv_2017}, have several orders of magnitude less labelled examples when compared to their image counterparts~\cite{deng_cvpr_2009, kuznetsova_ijcv_2020, sun_iccv_2017}.
As a result, training large models from scratch to high accuracy is extremely challenging.
To sidestep this issue, and enable more efficient training we initialise our video models from pretrained image models.
However, this raises several practical questions, specifically on how to initialise parameters not present or incompatible with image models.
We now discuss several effective strategies to initialise these large-scale video classification models.

\paragraph{Positional embeddings}
A positional embedding $\mathbf{p}$ is added to each input token (Eq.~\ref{eq:tokens}).
However, our video models have $n_t$ times more tokens than the pretrained image model. As a result, we initialise the positional embeddings by ``repeating'' them temporally from $\mathbb{R}^{n_w \cdot n_h \times d}$ to $\mathbb{R}^{n_t \cdot n_h \cdot n_w \times d}$.
Therefore, at initialisation, all tokens with the same spatial index have the same embedding which is then fine-tuned.

\paragraph{Embedding weights, $\mathbf{E}$}
When using the ``tubelet embedding'' tokenisation method (Sec.~\ref{sec:method_input_encoding}), the embedding filter $\mathbf{E}$ is a 3D tensor, compared to the 2D tensor in the pretrained model,  $\mathbf{E}_{\text{image}}$. A common approach for initialising 3D convolutional filters from 2D filters for video classification is to ``inflate'' them by replicating the filters along the temporal dimension and averaging them~\cite{carreira_cvpr_2017, feichtenhofer_neurips_2016} as
\begin{equation}
\mathbf{E} = \frac{1}{t}[\mathbf{E}_{\text{image}}, \ldots, \mathbf{E}_{\text{image}}, \ldots, \mathbf{E}_{\text{image}}].
\end{equation}
We consider an additional strategy, which we denote as ``central frame initialisation'', where $\mathbf{E}$ is initialised with zeroes along all temporal positions, except at the centre $\floor{\frac{t}{2}}$,
\begin{equation}
\mathbf{E} = [\mathbf{0}, \ldots,  \mathbf{E}_{\text{image}}, \ldots, \mathbf{0}].
\label{eq:central_frame_init}
\end{equation}
Therefore, the 3D convolutional filter effectively behaves like ``Uniform frame sampling'' (Sec.~\ref{sec:method_input_encoding}) at initialisation, while also enabling the model to learn to aggregate temporal information from multiple frames as training progresses.

\paragraph{Transformer weights for Model 3}
The transformer block in Model 3 (Fig.~\ref{fig:factorised_self_attention}) differs from the pretrained ViT model~\cite{dosovitskiy_iclr_2021}, in that it contains two multi-headed self attention (MSA) modules.
In this case, we initialise the spatial MSA module from the pretrained module, and initialise all weights of the temporal MSA with zeroes, such that Eq.~\ref{eq:factorised_self_att_temporal} behaves as a residual connection~\cite{he_cvpr_2016} at initialisation.

\section{Empirical evaluation}

We first present our experimental setup and implementation details in Sec.~\ref{sec:exp_setup}, before ablating various components of our model in Sec.~\ref{sec:exp_ablation}. We then present state-of-the-art results on five datasets in Sec.~\ref{sec:exp_sota_comparison}.

\subsection{Experimental Setup}\label{sec:exp_setup}

\paragraph{Network architecture and training}
Our backbone architecture follows that of ViT~\cite{dosovitskiy_iclr_2021} and BERT~\cite{devlin_naacl_2019}. We consider ViT-Base (ViT-B, $L$=$12$, $N_H$=$12$, $d$=$768$), ViT-Large (ViT-L, $L$=$24$, $N_H$=$16$, $d$=$1024$), and ViT-Huge (ViT-H, $L$=$32$, $N_H$=$16$, $d$=$1280$), where $L$ is the number of transformer layers, each with a self-attention block of $N_H$ heads and hidden dimension $d$. 
We also apply the same naming scheme to our models (e.g., ViViT-B/16x2 denotes a ViT-Base backbone with a tubelet size of $h \times w \times t = 16 \times  16 \times  2$). In all experiments, the tubelet height and width are equal. 
Note that smaller tubelet sizes correspond to more tokens at the input, and thus more computation.

We train our models using synchronous SGD and momentum, a cosine learning rate schedule and TPU-v3 accelerators. %
We initialise our models from a ViT image model trained either on ImageNet-21K~\cite{deng_cvpr_2009} (unless otherwise specified) or the larger JFT~\cite{sun_iccv_2017} dataset.
We implement our method using the Scenic library~\cite{dehghani2021scenic} and have released our code and models.

\paragraph{Datasets}

We evaluate the performance of our proposed models on a diverse set of video classification datasets:

\emph{Kinetics}~\cite{kay_arxiv_2017} consists of 10-second videos sampled at 25fps from YouTube.
We evaluate on both Kinetics 400 and 600, containing 400 and 600 classes respectively.
As these are dynamic datasets (videos may be removed from YouTube), we note our dataset sizes are approximately 267 000 and 446 000 respectively.

\emph{Epic Kitchens-100} consists of egocentric videos capturing daily kitchen activities spanning 100 hours and 90 000 clips~\cite{damen_arxiv_2020}.
We report results following the standard ``action recognition'' protocol.
Here, each video is labelled with a ``verb'' and a ``noun'' and we therefore predict both categories using a single network with two ``heads''. The top-scoring verb and action pair predicted by the network form an ``action'', and action accuracy is the primary metric.

\emph{Moments in Time}~\cite{monfort_pami_2019} consists of 800 000,  3-second YouTube clips that capture the gist of a dynamic scene involving animals, objects, people, or natural phenomena. 

\emph{Something-Something v2} (SSv2)~\cite{goyal_iccv_2017} contains 220 000 videos, with durations ranging from 2 to 6 seconds.
In contrast to the other datasets, the objects and backgrounds in the videos are consistent across different action classes, and this dataset thus places more emphasis on a model's ability to recognise fine-grained motion cues.

\begin{table}[t]
	\centering
	\caption{Comparison of input encoding methods using ViViT-B and spatio-temporal attention on Kinetics. 
	Further details in text.
	}
	\vspace{.1em}
	\begin{tabular}{lc}
		\toprule
		& Top-1 accuracy \\ \midrule
		Uniform frame sampling 		& 78.5          \\  %
		\midrule
		\textit{Tubelet embedding} \\
		Random initialisation~\cite{glorot_aistats_2010}  			& 73.2          \\ %
		Filter inflation~\cite{carreira_cvpr_2017}      				  & 77.6          \\ %
		Central frame          				& 79.2          \\ %
		\bottomrule
	\end{tabular}
	\vspace{-4mm}
	\label{tab:ablation_input_encoding}
\end{table}

\paragraph{Inference}
The input to our network is a video clip of 32 frames using a stride of 2, unless otherwise mentioned, similar to~\cite{feichtenhofer_iccv_2019, feichtenhofer_cvpr_2020}.
Following common practice, at inference time, we process multiple views of a longer video and average per-view logits to obtain the final result.
Unless otherwise specified, we use a total of 4 views per video (as this is sufficient to ``see'' the entire video clip across the various datasets), and ablate these and other design choices next.

\subsection{Ablation study}
\label{sec:exp_ablation}

\paragraph{Input encoding}
We first consider the effect of different input encoding methods (Sec.~\ref{sec:method_input_encoding}) using our unfactorised model (Model 1) and ViViT-B on Kinetics 400. 
As we pass 32-frame inputs to the network, sampling 8 frames and extracting tubelets of length $t = 4$ correspond to the same number of tokens in both cases.
Table~\ref{tab:ablation_input_encoding} shows that tubelet embedding initialised using the ``central frame'' method (Eq.~\ref{eq:central_frame_init}) performs well, outperforming the commonly-used ``filter inflation'' initialisation method~\cite{carreira_cvpr_2017, feichtenhofer_neurips_2016} by 1.6\%, and ``uniform frame sampling'' by 0.7\%.
We therefore use this encoding method for all subsequent experiments.

\begin{table}[tb]
	\centering
	\caption{Comparison of model architectures using ViViT-B as the backbone, and tubelet size of $16 \times 2$.
		We report Top-1 accuracy on Kinetics 400 (K400) and action accuracy on Epic Kitchens (EK).
		Runtime is during inference on a TPU-v3.
	}
	\scalebox{0.72}{
		\begin{tabular}{l M{0.09\linewidth} M{0.09\linewidth} M{0.13\linewidth} M{0.13\linewidth} M{0.13\linewidth}}
			\toprule
			& K400 & EK & FLOPs ($\times10^{9}$) & Params ($\times10^{6}$) & Runtime (ms) \\ \midrule
			Model 1: Spatio-temporal           			 &  80.0            &    43.1            &   455.2  & 88.9  &  58.9       \\			%
			Model 2: Fact. encoder        			&  78.8            &     43.7           &   284.4    & 115.1 &  17.4              \\   %
			Model 3: Fact. self-attention 		 &   77.4     &     39.1           &    372.3         &  117.3 &  31.7          \\  %
			Model 4: Fact. dot product    		 &   76.3       &       39.5         &    277.1    &  88.9 &  22.9         \\  %
			\midrule
			Model 2: Ave. pool baseline    & 75.8 &     38.8 	& 283.9 &	86.7 &  17.3  \\ %
			\bottomrule
		\end{tabular}
		\vspace{-5mm}
		\label{tab:ablation_model_variants}
	}
\end{table}

\begin{table}[tb]
	\centering
	\caption{The effect of varying the number of temporal transformers, $L_t$, in the Factorised encoder model (Model 2).
		We report the Top-1 accuracy on Kinetics 400.
		Note that $L_t = 0$ corresponds to the ``average pooling baseline''.
	}
	\begin{tabular}{lccccc}
	\toprule
	 $L_t$    			& 0 & 1 & 4 & 8 & 12 \\ \midrule
	 Top-1		& 75.8  & 78.6   & 78.8  & 78.8  & 78.9   \\ \bottomrule
	\end{tabular}
	\label{tab:ablation_temporal_transformers}
	\vspace{-\baselineskip}
\end{table}

\paragraph{Model variants}
We compare our proposed model variants (Sec.~\ref{sec:method_video_models}) across the Kinetics 400 and Epic Kitchens datasets, both in terms of accuracy and efficiency, in Tab.~\ref{tab:ablation_model_variants}.
In all cases, we use the ``Base'' backbone and tubelet size of $16 \times 2$.
Model 2 (``Factorised Encoder'') has an additional hyperparameter, the number of temporal transformers, $L_t$.
We set $L_t = 4$ for all experiments and show in Tab.~\ref{tab:ablation_temporal_transformers} that the model is not sensitive to this choice.

The unfactorised model (Model 1) performs the best on Kinetics 400.
However, it can also overfit on smaller datasets such as Epic Kitchens, where we find our ``Factorised Encoder'' (Model 2) to perform the best.
We also consider an additional baseline (last row), based on Model 2, where we do not use any temporal transformer, and simply average pool the frame-level representations from the spatial encoder before classifying.
This average pooling baseline performs the worst, and has a larger accuracy drop on Epic Kitchens, suggesting that this dataset requires more detailed modelling of temporal relations.

As described in Sec.~\ref{sec:method_video_models}, all factorised variants of our model use significantly fewer FLOPs than the unfactorised Model 1, as the attention is computed separately over spatial- and temporal-dimensions.
Model 4 adds no additional parameters to the unfactorised Model 1, and uses the least compute.
The temporal transformer encoder in Model 2 operates on only $n_t$ tokens, which is why there is a barely a change in compute and runtime over the average pooling baseline, even though it improves the accuracy substantially (3\% on Kinetics and 4.9\% on Epic Kitchens).
Finally, Model 3 requires more compute and parameters than the other factorised models, as its additional self-attention block means that it performs another query-, key-, value- and output-projection in each transformer layer~\cite{vaswani_neurips_2017}.

\begin{table}[t]
	\centering
	\caption{
		The effect of progressively adding regularisation (each row includes all methods above it) on Top-1 action accuracy on Epic Kitchens.
		We use a Factorised encoder model with tubelet size $16 \times 2$.
	}
	\small{  %
		\begin{tabular}{lcc}
			\toprule 
			& Top-1 accuracy  \\ 
			\midrule
			Random crop, flip, colour jitter           &      38.4   		\\   %
			+ Kinetics 400 initialisation 				 &      39.6  		\\  %
			+ Stochastic depth~\cite{huang_stochasticdepth_eccv_2016}       				 &      40.2 			\\  %
			+ Random augment~\cite{cubuk_arxiv_2019}             				&      41.1 	 \\   %
			+ Label smoothing~\cite{szegedy_cvpr_2016}             				  &      43.1      \\   %
			+ Mixup~\cite{zhang_mixup_iclr_2018}                       	 	  				&      43.7 	   \\   %
			\bottomrule
		\end{tabular}
	}
	\label{tab:regularisation_epic_kitchens}
	\vspace{-\baselineskip}
\end{table}

\begin{figure}[t]
	\centering
	\includegraphics[width=\linewidth]{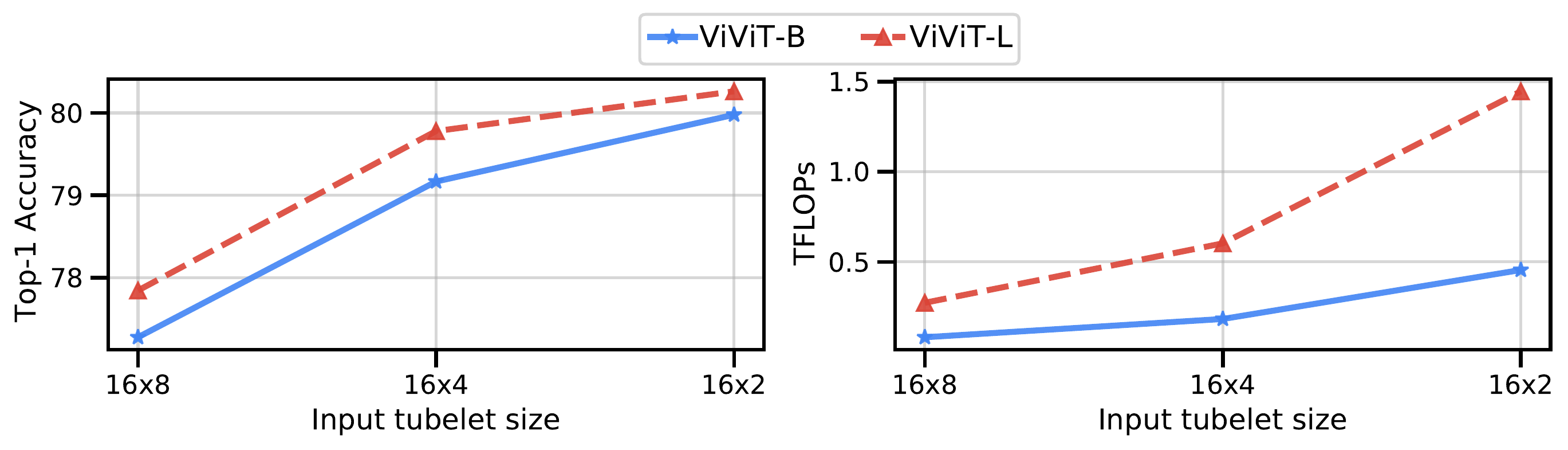}
	\footnotesize{
    	\begin{tabularx}{\linewidth}{YY}
    	    (a) Accuracy   &  (b) Compute
    	\end{tabularx}
	}
	\caption{The effect of the backbone architecture on (a) accuracy and (b) computation %
	on Kinetics 400, for the spatio-temporal attention model (Model 1).
	}
	\label{fig:ablation_backbone}
	\vspace{-0.5\baselineskip}
\end{figure}

\begin{figure}[t]
	\centering
	\includegraphics[width=\linewidth]{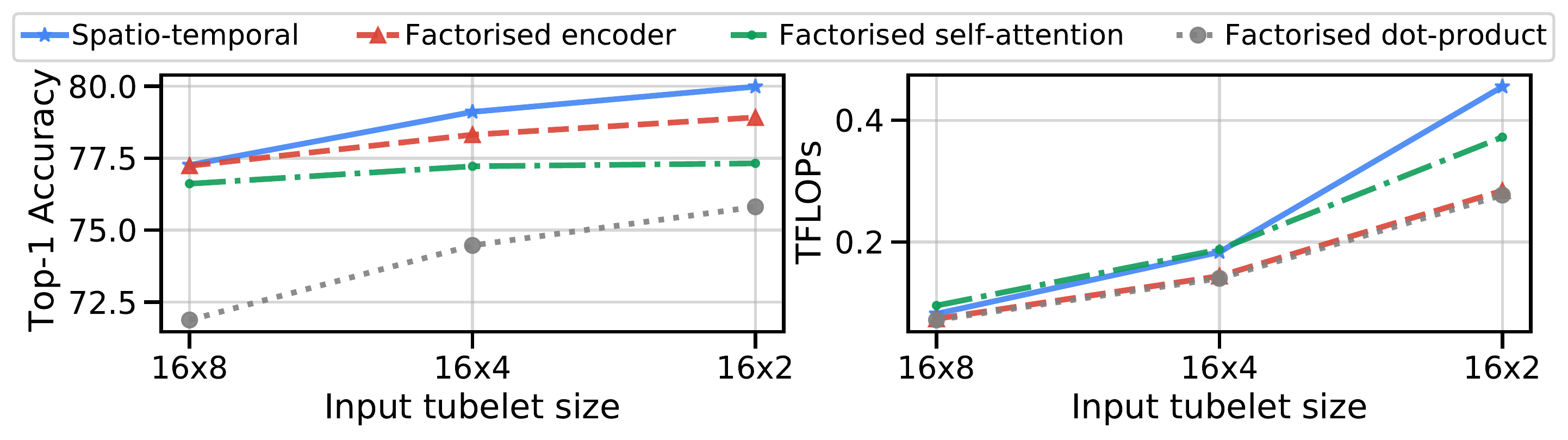}
	\\
	\footnotesize{
    	\begin{tabularx}{\linewidth}{YY}
    	    (a) Accuracy   &  (b) Compute
    	\end{tabularx}
	}
 	\vspace{-0.5\baselineskip}
	\caption{The effect of varying the number of temporal tokens on (a) accuracy and (b) computation %
	on Kinetics 400, for different variants of our model with a ViViT-B backbone.
	}
	\label{fig:ablation_temporal_tokens}
	\vspace{-\baselineskip}
\end{figure}

\paragraph{Model regularisation}
Pure-transformer architectures such as ViT~\cite{dosovitskiy_iclr_2021} are known to require large training datasets, and we observed overfitting on smaller datasets like Epic Kitchens and SSv2, even when using an ImageNet pretrained model.
In order to effectively train our models on such datasets, we employed several regularisation strategies that we ablate using our ``Factorised encoder'' model in Tab.~\ref{tab:regularisation_epic_kitchens}.
We note that these regularisers were originally proposed for training CNNs, and that~\cite{touvron_arxiv_2020} have recently explored them for training ViT for image classification.

Each row of Tab.~\ref{tab:regularisation_epic_kitchens} includes all the methods from the rows above it, and we observe progressive improvements from adding each regulariser.
Overall, we obtain a substantial overall improvement of 5.3\% on Epic Kitchens. %
We also achieve a similar improvement of 5\% %
on SSv2 by using all the regularisation in Tab.~\ref{tab:regularisation_epic_kitchens}.
Note that the Kinetics-pretrained models that we initialise from are from Tab.~\ref{tab:ablation_model_variants}, and that all Epic Kitchens models in Tab.~\ref{tab:ablation_model_variants} were trained with all the regularisers in Tab.~\ref{tab:regularisation_epic_kitchens}.
For larger datasets like Kinetics and Moments in Time, we do not use these additional regularisers (we use only the first row of Tab.~\ref{tab:regularisation_epic_kitchens}), as we obtain state-of-the-art results without them.
The appendix contains hyperparameter values and additional details for all regularisers.

\paragraph{Varying the backbone}
Figure~\ref{fig:ablation_backbone} compares the ViViT-B and ViViT-L backbones for the unfactorised spatio-temporal model. We observe consistent improvements in accuracy as the backbone capacity increases. As expected, the compute also grows as a function of the backbone size.

\begin{table}[t]
	\centering
	\caption{The effect of spatial resolution on the performance of ViViT-L/16x2 and spatio-temporal attention on Kinetics 400.}
	\vspace{0.1em}
	\small{
	\begin{tabular}{lccc}
		\toprule
		Crop size & 224         & 288         & 320        \\ \midrule
		Accuracy &   80.3          &  80.7            &   81.0         \\   %
		GFLOPs    &     1446        &   2919          &   3992         \\
		Runtime  &     58.9        &      147.6       &    238.8       \\
		\bottomrule
	\end{tabular}
	\vspace{-0.5\baselineskip}
	\label{tab:ablation_spatial_resolution}
	}
\end{table}

\paragraph{Varying the number of tokens}
We first analyse the performance as a function of the number of tokens along the temporal dimension %
in Fig.~\ref{fig:ablation_temporal_tokens}. 
We observe that using smaller input tubelet sizes (and therefore more tokens) leads to consistent accuracy improvements across all of our model architectures.
At the same time, computation in terms of FLOPs increases accordingly, and the unfactorised model (Model 1) is impacted the most.

We then vary the number of tokens fed into the model by increasing the spatial crop-size from the default of 224 to 320 in Tab.~\ref{tab:ablation_spatial_resolution}.
As expected, there is a consistent increase in both accuracy and computation. We note that when comparing to prior work we consistently obtain state-of-the-art results (Sec.~\ref{sec:exp_sota_comparison}) using a spatial resolution of 224, but we also highlight that further improvements can be obtained at higher spatial resolutions.

\begin{table*}[tb]
	\caption{Comparisons to state-of-the-art across multiple datasets. For ``views'', $x \times y$ denotes $x$ temporal crops and $y$ spatial crops.
	We report the TFLOPs to process all spatio-temporal views.
	``FE'' denotes our Factorised Encoder model.
	}
	\vspace{0.2\baselineskip}
	\begin{subtable}[t]{.40\linewidth}
		\centering
		\caption{Kinetics 400}
    	\setlength{\tabcolsep}{4pt} %
		\renewcommand*{\arraystretch}{1.11}  %
		
		\vspace{-0.3\baselineskip}
		\scriptsize{
			\begin{tabular}{lcccc}
				\toprule
				Method 																			 & Top 1                & Top 5         & Views 	& TFLOPs \\
				\midrule
				blVNet~\cite{fan_blvnet_neurips_2019}							  & 73.5 				  & 91.2  & --  & --\\ 
				STM~\cite{jiang_stm_iccv_2019}										& 73.7 					& 91.6	& -- & -- \\
				TEA~\cite{li_tea_cvpr_2020}												& 76.1					& 92.5 & $10 \times 3$ & 2.10 \\  %
				TSM-ResNeXt-101~\cite{lin_tsm_cvpr_2019}						  & 76.3				 & -- &  -- & -- \\
				I3D NL~\cite{wang_cvpr_2018}										 & 77.7                  & 93.3         		 &  $10 \times 3$ & 10.77     \\  %
				CorrNet-101~\cite{wang_corrnet_cvpr_2020}					  & 79.2				 & --			 		 & $10 \times 3$	& 6.72	 \\  %
				ip-CSN-152~\cite{tran_iccv_2019}									&  79.2					& 93.8				& $10 \times 3$	& 3.27	 \\  %
				LGD-3D R101~\cite{qiu_lgd_cvpr_2019}							&  79.4				    & 94.4			 		&  --	& --					\\  
				SlowFast R101-NL~\cite{feichtenhofer_iccv_2019}       		&  79.8                 &  93.9                   & $10 \times 3$    & 7.02   \\  %
				X3D-XXL~\cite{feichtenhofer_cvpr_2020}      					&  80.4					&  94.6			  		& $10 \times 3$   & 5.82   \\
				TimeSformer-L~\cite{bertasius_arxiv_2021}					  	& 80.7				& \textbf{94.7}					& $1 \times 3$ & 7.14 \\
				ViViT-L/16x2 FE 														  	 &  80.6 				& 92.7 & $1 \times 1$   & 3.98 \\  %
				ViViT-L/16x2 FE 															& \textbf{81.7} 	&  93.8  & $1 \times 3$ &  11.94 \\  %
				\midrule
				\multicolumn{4}{l}{\textit{Methods with large-scale pretraining}}                                \\ 
				ip-CSN-152~\cite{tran_iccv_2019} (IG~\cite{mahajan_eccv_2018}) 			  &  82.5					& 95.3			&  $10 \times 3$ & 3.27	  \\
				ViViT-L/16x2 FE (JFT) 														& 83.5 	&  94.3   & $1 \times 3$ &  11.94 \\  %
				ViViT-H/14x2 (JFT) 														&  \textbf{84.9} &  \textbf{95.8} 	& $4 \times 3$ & 47.77 \\  %
				\bottomrule
			\end{tabular}
		}
		\label{tab:sota_kinetics400}
	\end{subtable}
  	\hfill
  	\begin{subtable}[t]{.28\linewidth}
		\centering
  		\caption{Kinetics 600}
  		\setlength{\tabcolsep}{4pt} %
		\vspace{-0.3\baselineskip}
  		\scriptsize{
	  		\begin{tabular}{lcc}
	  			\toprule
	  			Method 																			 & Top 1                & Top 5      \\ %
	  			\midrule
	  			AttentionNAS~\cite{wang_nas_eccv_2020}						  &  79.8				   & 94.4  \\ %
	  			LGD-3D R101~\cite{qiu_lgd_cvpr_2019}							&  81.5				    & 95.6	\\ %
	  			SlowFast R101-NL~\cite{feichtenhofer_iccv_2019}       		&  81.8                     &  95.1   \\ %
	  			X3D-XL~\cite{feichtenhofer_cvpr_2020}      						 &  81.9					&  95.5		\\ %
	  			TimeSformer-L~\cite{bertasius_arxiv_2021}					  & 82.2				& \textbf{95.6}		\\ %
	  			ViViT-L/16x2 FE 														  	 &  \textbf{82.9} 	& 94.6  \\  %
	  			\midrule
	  			ViViT-L/16x2 FE (JFT) 													 & 84.3 & 94.9 \\ %
	  			ViViT-H/14x2 (JFT) 													& \textbf{85.8} & \textbf{96.5} \\ %
	  			\bottomrule
	  		\end{tabular}
	  		\label{tab:sota_kinetics600}
  		}
  		\vspace{0.35\baselineskip} %
  		\centering
  		\caption{Moments in Time}
  		\vspace{-0.3\baselineskip}
  		\setlength{\tabcolsep}{6pt} %
  		\scriptsize{
  			\begin{tabular}{lcc}
  				\toprule
  				& Top 1 & Top 5 \\ 
  				\midrule
  				TSN~\cite{wang_tsn_eccv_2016}				& 25.3		 &  50.1	\\
  				TRN~\cite{zhou_trn_eccv_2018}				 & 28.3		 &  53.4	 \\
  				I3D~\cite{carreira_cvpr_2017}					& 29.5		&  56.1		\\
  				blVNet~\cite{fan_blvnet_neurips_2019}	 &  31.4	 &  59.3 	 \\
  				AssembleNet-101~\cite{ryoo_iclr_2020} 	&  34.3     &  62.7     \\
  				\midrule
  				ViViT-L/16x2 FE 	& \textbf{38.5} & \textbf{64.1} \\  %
  				\bottomrule
  			\end{tabular}
  			\label{tab:sota_moments_in_time}
  		}
  	\end{subtable}
  	\hfill
  	\begin{subtable}[t]{.28\linewidth}
  		\caption{Epic Kitchens 100 Top 1 accuracy}
  		\setlength{\tabcolsep}{4pt} %
  		\vspace{-0.3\baselineskip}
		\scriptsize{
			\begin{tabular}{lccc}
				\toprule
				Method 													 & Action & Verb  & Noun  \\
				\midrule
				TSN~\cite{wang_tsn_eccv_2016}		 		&  33.2 & 60.2 & 46.0  \\
				TRN~\cite{zhou_trn_eccv_2018} 				& 35.3 & 65.9 & 45.4   \\
				TBN~\cite{kazakos_iccv_2019}				 & 36.7 & 66.0 & 47.2    \\
				TSM~\cite{lin_tsm_cvpr_2019} 				 & 38.3 & \textbf{67.9} & 49.0  \\
				SlowFast~\cite{feichtenhofer_iccv_2019}  & 38.5 & 65.6 & 50.0  \\
				\midrule
				ViViT-L/16x2 FE & \textbf{44.0} & 66.4 & \textbf{56.8} \\ %
				\bottomrule
			\end{tabular}
		}
		\label{tab:sota_epic_kitchens}
		\vspace{0.4\baselineskip}
		\centering
		\caption{Something-Something v2}
		\setlength{\tabcolsep}{6pt} %
		\scriptsize{
			\begin{tabular}{lcc}
				\toprule
				Method 												  & Top 1                & Top 5   \\
				\midrule
				TRN~\cite{zhou_trn_eccv_2018}										& 48.8 & 77.6 \\
				SlowFast~\cite{feichtenhofer_cvpr_2020,wu_multigrid_cvpr_2020}		& 61.7 & --    \\
				TimeSformer-HR~\cite{bertasius_arxiv_2021}					& 62.5 & -- \\
				TSM~\cite{lin_tsm_cvpr_2019}    		   							& 63.4 & 88.5 \\
				STM~\cite{jiang_stm_iccv_2019}				   						& 64.2 & 89.8 \\
				TEA~\cite{li_tea_cvpr_2020}						   					& 65.1 & -- \\
				blVNet~\cite{fan_blvnet_neurips_2019}  		 						& 65.2  & \textbf{90.3} \\
				\midrule
				ViVIT-L/16x2 FE		& \textbf{65.9}  & 89.9 \\ %
				\bottomrule
			\end{tabular}
			\label{tab:sota_ssv2}
		}
	\end{subtable}
	\vspace{-\baselineskip}
\end{table*}

\begin{figure}[tb]
	\centering
	\includegraphics[width=0.95\linewidth]{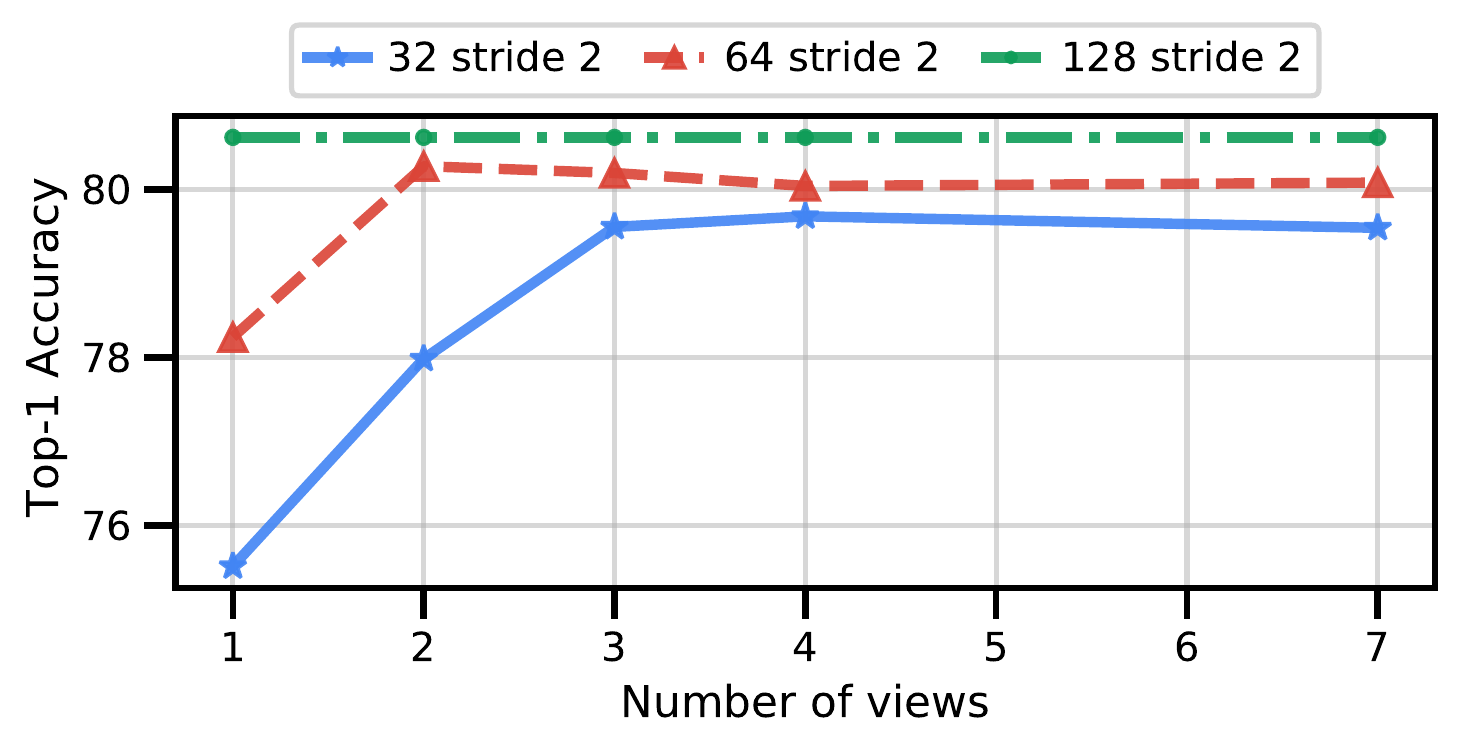}
	\caption{
		The effect of varying the number of frames input to the network and increasing the number of tokens proportionally.
		We use ViViT-L/16x2 Factorised Encoder on Kinetics 400.
		A Kinetics video contains 250 frames (10 seconds sampled at 25 fps) and the accuracy for each model saturates once the number of equidistant temporal views is sufficient to ``see'' the whole video clip.
		Observe how models processing more frames (and thus more tokens) achieve higher single- and multi-view accuracy.
	}
	\label{fig:ablation_num_frames}
	\vspace{-\baselineskip}
\end{figure}
\paragraph{Varying the number of input frames}

In our experiments so far, we have kept the number of input frames fixed at 32.
We now increase the number of frames input to the model, thereby increasing the number of tokens proportionally.

Figure~\ref{fig:ablation_num_frames} shows that as we increase the number of frames input to the network, the accuracy from processing a single view increases, since the network incorporates longer temporal context.
However, common practice on datasets such as Kinetics~\cite{feichtenhofer_iccv_2019, wang_cvpr_2018, li_tea_cvpr_2020} is to average results over multiple, shorter ``views'' of the same video clip.
Figure~\ref{fig:ablation_num_frames} also shows that the accuracy saturates once the number of views is sufficient to cover the whole video.
As a Kinetics video consists of 250 frames, and we sample frames with a stride of 2, our model which processes 128 frames requires just a single view to ``see'' the whole video and achieve its maximum accuarcy.

Note that we used  ViViT-L/16x2 Factorised Encoder (Model 2) here.
As this model is more efficient it can process more tokens, compared to the unfactorised Model 1 which runs out of memory after 48 frames using tubelet length $t = 2$ and a ``Large'' backbone.
Models processing more frames (and thus more tokens) consistently achieve higher single- and multi-view accuracy, in line with our observations in previous experiments (Tab.~\ref{tab:ablation_spatial_resolution}, Fig.~\ref{fig:ablation_temporal_tokens}).
Moroever, observe that by processing more frames (and thus more tokens) with Model 2, we are able to achieve higher accuracy than Model 1 (with fewer total FLOPs as well).

Finally, we observed that for Model 2, the number of FLOPs effectively increases linearly with the number of input frames as the overall computation is dominated by the initial Spatial Transformer.
As a result, the total number of FLOPs for the number of temporal views required to achieve maximum accuracy is constant across the models.
In other words, ViViT-L/16x2 FE with 32 frames requires 995.3 GFLOPs per view, and 4 views to saturate multi-view accuracy.
The 128-frame model requires 3980.4 GFLOPs but only a single view.
As shown by Fig.~\ref{fig:ablation_num_frames}, the latter model achieves the highest accuracy.

\subsection{Comparison to state-of-the-art}\label{sec:exp_sota_comparison}
Based on our ablation studies in the previous section, we compare to the current state-of-the-art using two of our model variants.
We primarily use our Factorised Encoder model (Model 2), as it can process more tokens than Model 1 to achieve higher accuracy.

\vspace{-0\baselineskip}
\paragraph{Kinetics}
Tables~\ref{tab:sota_kinetics400} and ~\ref{tab:sota_kinetics600} show that our spatio-temporal attention models outperform the state-of-the-art on Kinetics 400 and 600 respectively.
Following standard practice, we take 3 spatial crops (left, centre and right)~\cite{feichtenhofer_iccv_2019,feichtenhofer_cvpr_2020,tran_iccv_2019,wang_cvpr_2018} for each temporal view, and notably, we require significantly fewer views than previous CNN-based methods.

We surpass the previous CNN-based state-of-the-art using ViViT-L/16x2 Factorised Encoder (FE) pretrained on ImageNet, and also outperform~\cite{bertasius_arxiv_2021} who concurrently proposed a pure-transformer architecture.
Moreover, by initialising our backbones from models pretrained on the larger JFT dataset~\cite{sun_iccv_2017}, we obtain further improvements. Although these models are not directly comparable to previous work, we do also outperform~\cite{tran_iccv_2019} who pretrained on the large-scale, Instagram dataset~\cite{mahajan_eccv_2018}. Our best model uses a ViViT-H backbone pretrained on JFT and significantly advances the best reported results on Kinetics 400 and 600 to 84.9\% and 85.8\%, respectively.

\vspace{-0\baselineskip}
\paragraph{Moments in Time}
We surpass the state-of-the-art by a significant margin as shown in Tab.~\ref{tab:sota_moments_in_time}. 
We note that the videos in this dataset are diverse and contain significant label  noise, making this task challenging and leading to lower accuracies than on other datasets.

\vspace{-0\baselineskip}
\paragraph{Epic Kitchens 100}
Table~\ref{tab:sota_epic_kitchens} shows that our Factorised Encoder model outperforms previous methods by a significant margin. In addition, our model obtains substantial improvements for Top-1 accuracy of ``noun'' classes, and the only method which achieves higher ``verb'' accuracy used optical flow as an additional input modality~\cite{lin_tsm_cvpr_2019, price_arxiv_2019}. 
Furthermore, all variants of our model presented in Tab.~\ref{tab:ablation_model_variants} outperformed the existing state-of-the-art on  action accuracy.
We note that we use the same model to predict verbs and nouns using two separate ``heads'', and for simplicity, we do not use separate loss weights for each head. 

\paragraph{Something-Something v2 (SSv2)}

Finally, Tab.~\ref{tab:sota_ssv2} shows that we achieve state-of-the-art Top-1 accuracy with our Factorised encoder model (Model 2), albeit with a smaller margin compared to previous methods.
Notably, our Factorised encoder model significantly outperforms the concurrent TimeSformer~\cite{bertasius_arxiv_2021} method by 2.9\%, which also proposes a pure-transformer model, but does not consider our Factorised encoder variant or our additional regularisation.

SSv2 differs from other datasets in that the backgrounds and objects are quite similar across different classes, meaning that recognising fine-grained motion patterns is necessary to distinguish classes from each other.
Our results suggest that capturing these fine-grained motions is an area of improvement and future work for our model.
We also note an inverse correlation between the relative performance of previous methods on SSv2 (Tab.~\ref{tab:sota_ssv2}) and Kinetics~(Tab.~\ref{tab:sota_kinetics400})
suggesting that these two datasets evaluate complementary characteristics of a model.

\section{Conclusion and Future Work}

We have presented four pure-transformer models for video classification, with different accuracy and efficiency profiles, achieving state-of-the-art results across five popular datasets.
Furthermore, we have shown how to effectively regularise such high-capacity models for training on smaller datasets and thoroughly ablated our main design choices.
Future work is to remove our dependence on image-pretrained models.
Finally, going beyond video classification towards more complex tasks is a clear next step.

{\small
\bibliographystyle{ieee_fullname}
\bibliography{bibliography}
}

\clearpage
\appendix

\section*{Appendix}
\section{Additional experimental details}

In this appendix, we provide additional experimental details.
Section~\ref{sec:regulariser_details} provides additional details about the regularisers we used and Sec.~\ref{sec:hyperparameters} details the training hyperparamters used for our experiments.

\begin{table*}[tb]
\caption{Training hyperparamters for experiments in the main paper.  ``--'' indicates that the regularisation method was not used at all. 
Values which are constant across all columns are listed once.
Datasets are denoted as follows: K400: Kinetics 400. K600: Kinetics 600. MiT: Moments in Time. EK: Epic Kitchens. SSv2: Something-Something v2.}
\begin{tabularx}{\linewidth}{lYYYYY}
\toprule
                              & K400 & K600 & MiT & EK & SSv2 \\ \midrule
\multicolumn{6}{l}{\textit{Optimisation}}                                                                                 \\
Optimiser               & \multicolumn{5}{c}{Synchronous SGD}                                                               \\
Momentum					& \multicolumn{5}{c}{0.9} \\
Batch size 						& \multicolumn{5}{c}{64} \\
Learning rate schedule	& \multicolumn{5}{c}{cosine with linear warmup} \\
Linear warmup epochs	& \multicolumn{5}{c}{2.5} \\
Base learning rate			& 0.1 & 0.1 & 0.25 & 0.5 & 0.5 \\
Epochs	& 30 & 30 & 10 & 50 & 35 \\
\midrule
\multicolumn{6}{l}{\textit{Data augmentation}} \\
Random crop probability & \multicolumn{5}{c}{1.0} \\
Random flip probability & \multicolumn{5}{c}{0.5} \\
Scale jitter probability	& \multicolumn{5}{c}{1.0} \\
Maximum scale			  & \multicolumn{5}{c}{1.33} \\
Minimum scale			  & \multicolumn{5}{c}{0.9} \\
Colour jitter probability  & 0.8 & 0.8 & 0.8 & -- & -- \\
Rand augment number of layers~\cite{cubuk_arxiv_2019}		 & -- & -- & -- & 2 & 2 \\
Rand augment magnitude~\cite{cubuk_arxiv_2019} & -- & -- & -- & 15 & 20 \\
\midrule
\multicolumn{6}{l}{\textit{Other regularisation}} \\
Stochastic droplayer rate, $p_{\text{drop}}$~\cite{huang_stochasticdepth_eccv_2016} & -- & -- & -- & 0.2 &  0.3 \\
Label smoothing $\lambda$~\cite{szegedy_cvpr_2016}				& -- & -- & -- & 0.2 & 0.3 \\
Mixup $\alpha$~\cite{zhang_mixup_iclr_2018}				 & -- & -- & -- & 0.1 & 0.3 \\
\bottomrule
\end{tabularx}
\label{tab:training_hyperparameters}
\end{table*}

\subsection{Further details about regularisers}
\label{sec:regulariser_details} 

In this section, we provide additional details and list the hyperparameters of the additional regularisers that we employed in Tab.~\ref{tab:regularisation_epic_kitchens}.
Hyperparameter values for all our experiments are listed in Tab.~\ref{tab:training_hyperparameters}.

\paragraph{Stochastic depth}

Stochastic depth regularisation was originally proposed for training very deep residual networks~\cite{huang_stochasticdepth_eccv_2016}.
Intuitively, the outputs of a layer, $\ell$, are ``dropped out'' with probability, $p_{\text{drop}}(\ell)$ during training, by setting the output of the layer to be equal to its input.

Following~\cite{huang_stochasticdepth_eccv_2016}, we linearly increase the probability of dropping a layer according to its depth within the network, 
\begin{equation}
p_{\text{drop}}(\ell) = \frac{\ell}{L} p_{\text{drop}},
\end{equation}
where $\ell$ is the index of the layer in the network, and $L$ is the total number of layers.

\paragraph{Random augment}
Random augment~\cite{cubuk_arxiv_2019} randomly applies data augmentation transformations sequentially to an input example.
We follow the public implementation\footnote{\url{https://github.com/tensorflow/models/blob/master/official/vision/beta/ops/augment.py}}, but modify the data augmentation operations to be temporally consistent throughout the video (in other words, the same transformation is applied on each frame of the video).

The authors define two hyperparameters for Random augment, ``number of layers'' , the number of augmentation transformations to apply sequentially to a video and ``magnitude'',  the strength of the transformation that is shared across all augmentation operations.
Our values for these parameters are shown in Tab.~\ref{tab:training_hyperparameters}.

\paragraph{Label smoothing}
Label smoothing was proposed by~\cite{szegedy_cvpr_2016} originally to regularise training Inception-v3.
Concretely, the label distribution used during training, $\tilde{y}$, is a mixture of the one-hot ground-truth label, $y$,  and a uniform distribution, $u$, to encourage the network to produce less confident predictions during training:
\begin{equation}
\tilde{y} = (1 - \lambda) y + \lambda u.
\end{equation}
There is therefore one scalar hyperparamter, $\lambda \in [0, 1]$.

\paragraph{Mixup}
Mixup~\cite{zhang_mixup_iclr_2018} constructs virtual training examples which are a convex combination of pairs of training examples and their labels.
Concretely, given $(x_i, y_i)$ and $(x_j, y_j)$ where $x_i$ denotes an input vector and $y_i$ a one-hot input label, mixup constructs the virtual training example,
\begin{align}
\tilde{x} &= \lambda x_i + (1 - \lambda) x_j \nonumber \\
\tilde{y} &= \lambda y_i + (1 - \lambda) y_j.
\end{align}
$\lambda \in [0, 1]$, and is sampled from a Beta distribution, $\text{Beta}(\alpha, \alpha)$.
Our choice of the hyperparameter $\alpha$ is detailed in Tab.~\ref{tab:training_hyperparameters}.

\subsection{Training hyperparameters}
\label{sec:hyperparameters}

Table~\ref{tab:training_hyperparameters} details the hyperparamters for all of our experiments.
We use synchronous SGD with momentum, a cosine learning rate schedule with linear warmup, and a batch size of 64 for all experiments.
As aforementioned, we only employed additional regularisation when training on the smaller Epic Kitchens and Something-Something v2 datasets.

\end{document}